%% file: main.tex
\documentclass[10pt,twocolumn,letterpaper]{article}

\usepackage[pagenumbers]{cvpr} 

\input{preamble}

\input{macros}
\input{math_commands}

\definecolor{cvprblue}{rgb}{0.21,0.49,0.74}
\usepackage[pagebackref,breaklinks,colorlinks,allcolors=cvprblue]{hyperref}

\usepackage{marvosym}
\usepackage{url}

\def\paperID{*****} 
\def\confName{CVPR}
\def\confYear{2025}

\title{FramePainter: Endowing Interactive Image Editing with \\
Video Diffusion Priors}

\author{Yabo Zhang\textsuperscript{1} \ \ \
Xinpeng Zhou\textsuperscript{1}\ \ \
Yihan Zeng\textsuperscript{2}\ \ \ 
Hang Xu\textsuperscript{2}\ \ \
Hui Li\textsuperscript{1}\ \ \ 
Wangmeng Zuo\textsuperscript{1, \Letter}\\
\textsuperscript{1}Harbin Institute of Technology \quad
\textsuperscript{2}Huawei Noah’s Ark Lab \\
}

\begin{document}

\input{figText/introduction}
\input{text/0_abstract}
\input{text/1_introduction}
\input{text/2_related_work}
\input{text/3_method}
\input{text/4_experiments}
\input{text/5_conclusion}
{
    \small
    \bibliographystyle{ieeenat_fullname}
    \bibliography{main}
}

\input{text/6_appendix}

\end{document}

%% file: preamble.tex
\usepackage{bm}
\usepackage{makecell}

%% file: macros.tex
\newcommand{\ignorethis}[1]{}

\makeatletter
\DeclareRobustCommand\onedot{\futurelet\@let@token\@onedot}
\def\@onedot{\ifx\@let@token.\else.\null\fi\xspace}

\def\eg{\emph{e.g}\onedot} 
\def\ie{\emph{i.e}\onedot}

\makeatother

\newrobustcmd{\B}{\bfseries}
\newcommand*{\rom}[1]{\expandafter\romannumeral #1}

\definecolor{mydarkblue}{rgb}{0,0.08,1}
\definecolor{mydarkgreen}{rgb}{0.02,0.6,0.02}
\definecolor{mydarkred}{rgb}{0.8,0.02,0.02}
\definecolor{mydarkorange}{rgb}{0.40,0.2,0.02}
\definecolor{mypurple}{RGB}{111,0,255}
\definecolor{myred}{rgb}{1.0,0.0,0.0}
\definecolor{mygold}{rgb}{0.75,0.6,0.12}
\definecolor{myblue}{rgb}{0,0.2,0.8}
\definecolor{mydarkgray}{rgb}{0.66,0.66,0.66}



%% file: math_commands.tex
\def\eqref#1{equation~\ref{#1}}

\def\1{\bm{1}}



%% file: figText/introduction.tex
\twocolumn[{
\maketitle
\begin{center}
\vspace{-1.6em}
\includegraphics[width=0.99\linewidth]{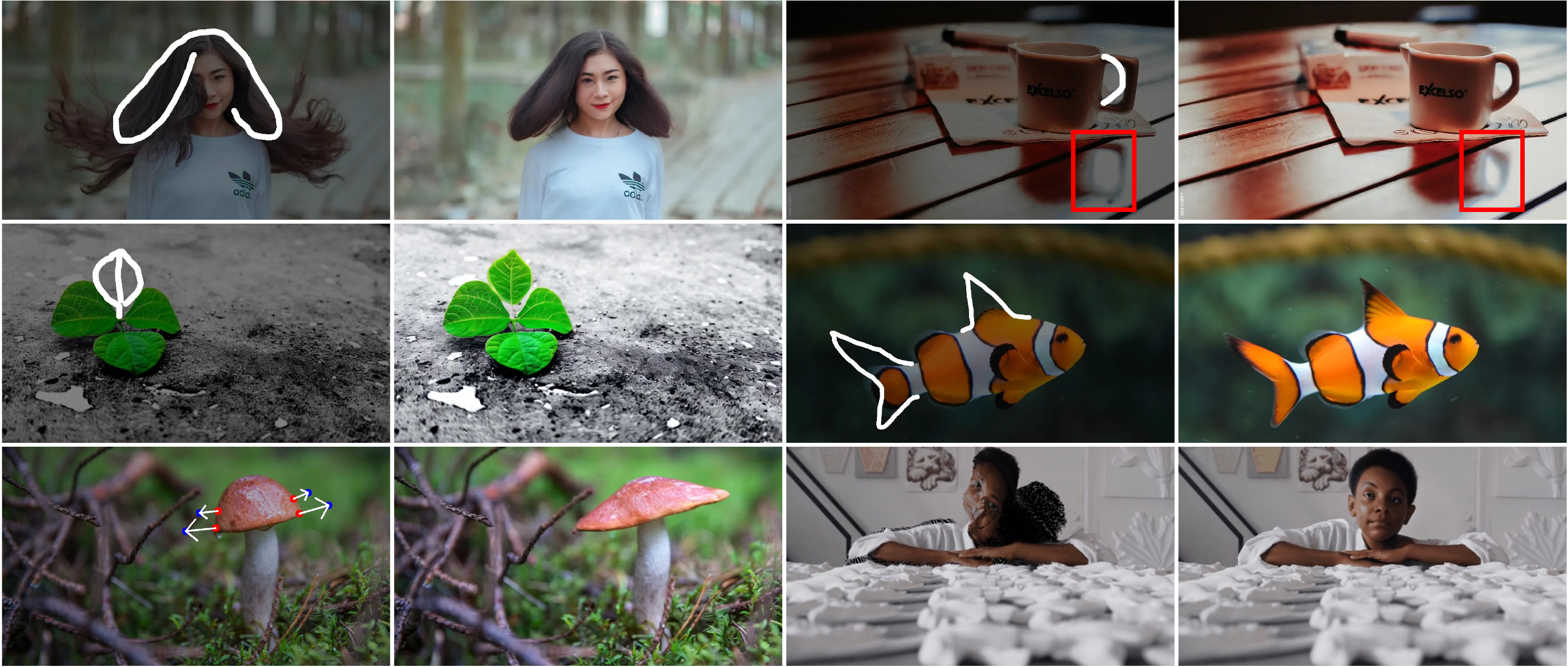}
\end{center}
\vspace{-1.2em}
\captionsetup{type=figure}
\captionof{figure}{
\textbf{Examples of FramePainter}.
FramePainter allows users to manipulate images through intuitive visual instructions like drawing sketches, clicking points, and dragging regions.
Benefiting from powerful video diffusion priors, it not only enables intuitive and plausible edits in common scenarios (\eg, adjust the
reflection of the cup in \textcolor{red}{red box}), but also exhibits exceptional generalization in out-of-domain cases, \eg, transform the clownfish into shark-like shape.
}\label{fig:teaser}
\vspace{1.35em}
}]

%% file: text/0_abstract.tex
\begin{abstract}
Interactive image editing allows users to modify images through visual interaction operations such as drawing, clicking, and dragging. 
Existing methods construct such supervision signals from videos, as they capture how objects change with various physical interactions.
However, these models are usually built upon text-to-image diffusion models, so necessitate (i) massive training samples and (ii) an additional reference encoder to learn real-world dynamics and visual consistency.
In this paper, we reformulate this task as an image-to-video generation problem, so that inherit powerful video diffusion priors to reduce training costs and ensure temporal consistency.
Specifically, we introduce FramePainter as an efficient instantiation of this formulation. 
Initialized with Stable Video Diffusion, it only uses a lightweight sparse control encoder to inject editing signals.
Considering the limitations of temporal attention in handling large motion between two frames, we further propose matching attention to enlarge the receptive field while encouraging dense correspondence between edited and source image tokens.
We highlight the effectiveness and efficiency of FramePainter across various of editing signals: it domainantly outperforms previous state-of-the-art methods with far less training data, achieving highly seamless and coherent editing of images, \eg, automatically adjust the reflection of the cup.
Moreover, FramePainter also exhibits exceptional generalization in scenarios not present in real-world videos, \eg, transform the clownfish into shark-like shape.
Our code will be available at \url{https://github.com/YBYBZhang/FramePainter}.
\end{abstract}

%% file: text/1_introduction.tex
\section{Introduction}
Diffusion models have achieved remarkable success in generating exceptional and photorealistic images from natural language descriptions~\cite{ramesh2022hierarchical,saharia2022photorealistic,rombach2022high,podell2023sdxl}.
Compared to generating images from scratch, the users prefer utilizing these models to edit captured photographs or designed images.
While text-guided image editing demonstrates promising potential~\cite{mokady2023nulltextinversion, meng2021sdedit, kawar2023imagic,cao2023masactrl, hertz2022prompt2prompt,brooks2023instructpix2pix}, it is constrained by the ambiguity of language instructions and the lack of precise spatial control, \eg, failing to accurately adjust the shape, position, or posture of a human.
In contrast, interactive image editing~\cite{Liu2024DragYN,Cui2024StableDragSD,Pan2023DragYG,Shi2024LightningDragLF,Alzayer2024MagicFS} offers a more flexible and precise solution, which supports more intuitive operations like drawing sketches, clicking points, and dragging regions.

Most existing approaches~\cite{Shi2024LightningDragLF,Alzayer2024MagicFS,mou2023dragondiffusion,Shi2023DragDiffusionHD} treat interactive image editing as an image-to-image generation task and leverage pre-trained text-to-image diffusion models~\cite{ramesh2022hierarchical,saharia2022photorealistic,rombach2022high,podell2023sdxl} as foundation models.
During finetuning, they~\cite{Shi2024LightningDragLF,Alzayer2024MagicFS} usually construct source and target image pairs from real-world videos, which provide sufficient observations of how objects change during physical interactions, \eg, a man raising his head.
However, in the absence of priors about real-world dynamics, these methods necessitate enormous training samples to achieve plausible and precise control over images.
Additionally, they require additional reference branches (\eg, the Reference U-Net~\cite{hu2024animate} or IP-Adapter~\cite{ye2023ip}) to maintain appearance consistency, further increasing model complexity and training costs.

In this paper, we reformulate interactive image editing as an image-to-video generation problem: 
the source image acts as the first frame, and an editing signal directs the generation of a video comprising the source and target images.
Our novel formulation enables the image editing models to leverage real-world dynamic priors from pre-trained video diffusion models~\cite{blattmann2023stable}, thereby potentially reducing both data and computational requirements compared to existing approaches.
Moreover, leveraging the features of the first frame for identity preservation, it eliminates the need for additional reference encoders and simplifies the model architecture.

Within this formulation, we propose \textit{FramePainter} to achieve flexibly and precisely interactive image editing using powerful video diffusion priors.
Specifically, FramePainter is initialized with Stable Video Diffusion (SVD)~\cite{blattmann2023stable} and uses a lightweight sparse control encoder to inject an editing signal (\eg, a sketch image) into the U-Net.
Since the temporal attention of SVD struggles with handling large motion between two images, we further introduce \textit{matching attention} to encourage dense correspondence between edited and source image tokens.
In particular, matching attention extends spatial attention along the temporal axis to enlarge the receptive field between images, and is implemented as an auxiliary branch to complement spatial attention.
To capture fine-grained visual details more precisely, we optimize matching attention by aligning its attention weights with tracking results from CoTracker-v3~\cite{karaev2024cotracker3}.
During inference, matching attention does not require tracking results as input, but can accurately query the corresponding source image token for each edited image token (refer to Fig.~\ref{fig:main_ablation_attn_vis} for visualization).

To verify the effectiveness of FramePainter, we construct thousands of image pairs from high-quality videos and split them into training and evaluation datasets.
Concretely, each image pair is randomly sampled and will be saved if their optical flow magnitude is large enough on local area (\ie, significant object movement).
Given two appropriate images, we extract corresponding editing signals with their optical flow or trackings, \eg, sketch images and dragging points.
Through extensive experiments on a wide range of editing signals, FramePainter not only significantly outperforms training-free methods in editing plausibility and visual consistency, but also surpasses training-based methods with substantially lower training costs.
Moreover, FramePainter demonstrates remarkable generalization in scenarios that are absent in real-world videos, such as transforming a clownfish into a shark-like shape.

Our core contributions are summarized as:
\begin{itemize}
    \item We reformulate interactive image editing as an image-to-video generation task, and introduce FramePainter to facilitate flexible and precise image manipulation using powerful video diffusion priors.
    \item To capture fine-grained visual appearance more precisely, we propose matching attention to encourage dense correspondence between edited and source image tokens.
    \item The experiments demonstrate that FramePainter achieves superior performance across various editing signals with far less training costs, while also showcasing exceptional generalization to unseen scenarios in real-world videos.
\end{itemize}

%% file: text/2_related_work.tex
\input{figText/method}
\section{Related Work}
\noindent \textbf{Image and Video Diffusion Models.}
Diffusion models have become the de facto standard in generative modeling and drive significant advancements in both image~\cite{ramesh2022hierarchical,nichol2022glide,saharia2022photorealistic,rombach2022high,podell2023sdxl,zhang2023adding,balaji2022ediffi} and video generation~\cite{chen2023videocrafter1,zhou2022magicvideo,he2022latent,wang2023modelscope,luo2023videofusion,ma2023follow,khachatryan2023text2video,zhang2023controlvideo,wu2023tune,ge2023preserve,bar2024lumiere,blattmann2023stable,xing2023dynamicrafter,zhang2023i2vgen,zeng2023make}.
In the field image synthesis, powerful foundational models such as Stable Diffusion~\cite{rombach2022high,podell2023sdxl} demonstrate unprecedented capabilities in producing realistic and diverse images and are thus widely applied to downstream tasks like text-guided image editing~\cite{tumanyan2023pnp,mokady2022prompt2prompt,mokady2023nulltext,parmar2023zero,instructpix2pix,cao2023masactrl} and controllable generation~\cite{zhang2023controlnet,mou2024t2iadapter,zhao2024uni,lv2024place,liu2025smartcontrol,wei2023elite,huang2023dreamcontrol,zhang2024videoelevator,wei2024masterweaver}.
Built upon pre-trained image diffusion models, video diffusion models~\cite{chen2023videocrafter1,zhou2022magicvideo,he2022latent,wang2023modelscope,wu2023tune,ge2023preserve,bar2024lumiere,blattmann2023stable,xing2023dynamicrafter,zhang2023i2vgen,zeng2023make} inherit their capability to generate high-quality video frames.
By introducing temporal modules and training on large-scale videos, they excel in understanding and recreating real-world dynamic processes by capturing temporal consistency and natural motion patterns.
However, due to the limited receptive field, it is challenging for temporal modules to deal with large movements in two frames, which can be mitigated by our proposed matching attention.

\noindent \textbf{Interactive Image Editing.}
The breakthroughs of diffusion models in text-to-image generation have significantly propelled advancements in the domain of image editing.
Among them, text-guided image editing~\cite{mokady2023nulltextinversion, meng2021sdedit, kawar2023imagic,cao2023masactrl, hertz2022prompt2prompt,brooks2023instructpix2pix,tumanyan2023pnp,mokady2023nulltext,parmar2023zero} are limited by the inherent ambiguity of language instructions and the absence of precise spatial control, while interactive image editing~\cite{Liu2024DragYN,Cui2024StableDragSD,Pan2023DragYG,Shi2024LightningDragLF,Alzayer2024MagicFS,mou2023dragondiffusion,nie2023blessing,Luo2023ReadoutGL} allows users to more flexibly and precisely manipulate images through intuitive instructions, such as drawing sketches, clicking points, and dragging regions~\cite{Alzayer2024MagicFS}.
An earlier work DragGAN~\cite{Pan2023DragYG} obtains promising point-based editing with motion supervision and point tracking but is limited, but is constrained by the inherent capacity of StyleGAN. 
DragDiffusion~\cite{Shi2023DragDiffusionHD} and DragonDiffusion~\cite{mou2023dragondiffusion} achieve open-domain point-based editing with the power of text-to-image diffusion models. 
Magic Fixup~\cite{Alzayer2024MagicFS} proposes a novel way to modify images, where an image is first coarsely edited by users and then transformed to realistic image through the network.
Existing works are tailored to a single type of editing signal and primarily focus on clicking points or dragging regions.
By contrast, FramePainter is capable of various editing signals and introduces drawing sketch as a more intuitive and convenient editing signal.

\noindent \textbf{Video Diffusion Priors for Generative Tasks.}
Video diffusion models~\cite{chen2023videocrafter1,zhou2022magicvideo,he2022latent,wang2023modelscope,ge2023preserve,bar2024lumiere,blattmann2023stable,xing2023dynamicrafter,zhang2023i2vgen,zeng2023make} possess stronger priors than image diffusion models when modeling real-world dynamic processes, particularly in capturing temporal consistency and plausible physical interactions.
While most existing works utilize these models for video-related downstream tasks~\cite{park2024spectral,tu2024motioneditor,ma2024follow,guo2025sparsectrl,he2024cameractrl,wei2024dreamvideo,jiang2024videobooth}, a few studies have explored the potential of directly learning from videos to enhance image-related tasks.
For example, Anydoor~\cite{Chen2023AnyDoorZO} diversifies object poses and viewpoints by borrowing knowledge from videos.
MagicFixup~\cite{Alzayer2024MagicFS} and LightningDrag~\cite{Shi2024LightningDragLF} select appropriate video frames as supervision signals of image editing task.
However, since these models are built upon image diffusion models, they requires extensive training data and additional
reference encoders to learn video knowledge from scratch.
Differently, FramePainter pioneers the integration of video diffusion priors into image editing task, significantly reducing training costs and simplifying the model architecture.

%% file: figText/method.tex
\begin{figure*}[t!]
   \begin{center}
   \includegraphics[width=.95\linewidth]{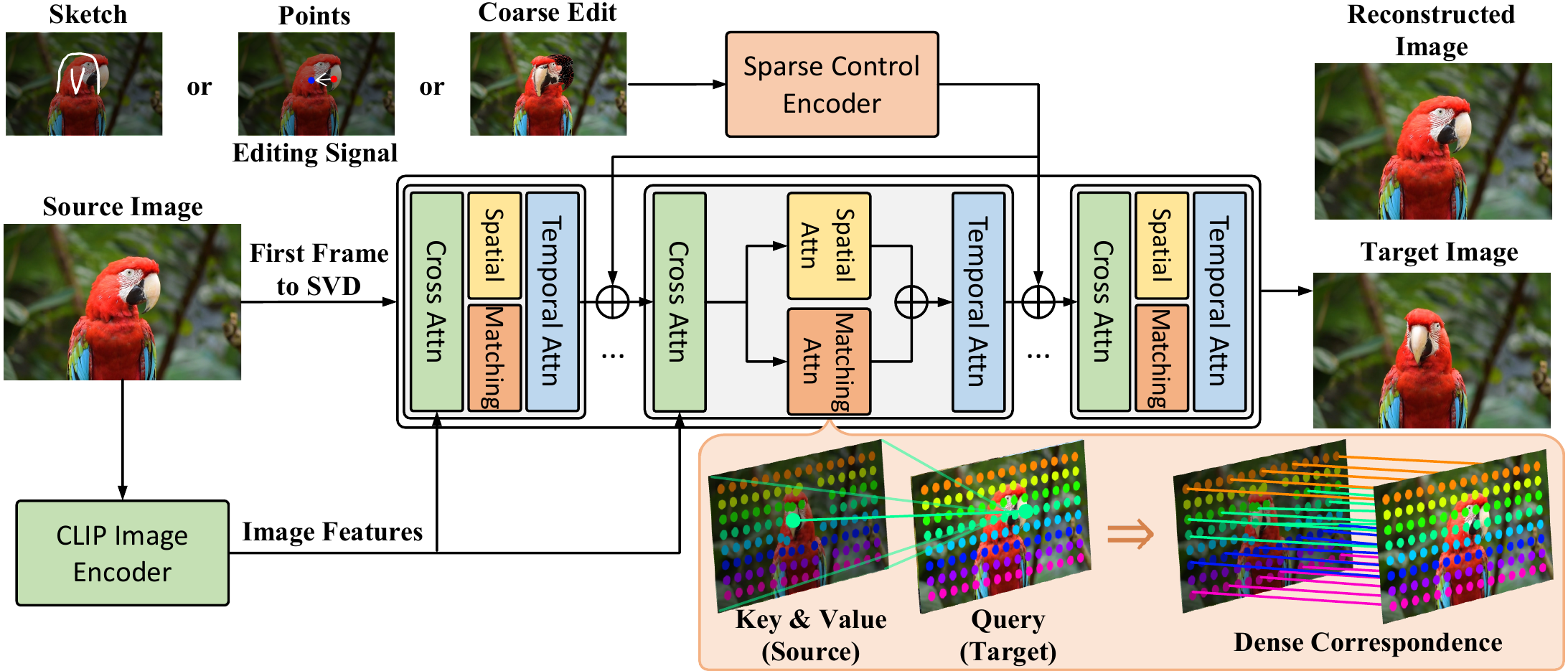}
   \end{center}
   \caption{
   \textbf{Overview of FramePainter.}
   Reformulating image editing as an image-to-video generation task, FramePainter takes a source image and an editing instruction as the first frame and control guidance, and produces a two-frame video comprising of reconstructed and target images.
    To improve visual consistency of two images involving large motion, matching attention is proposed to enlarge the receptive field and encourage dense correspondence between target and source image tokens.
   } 
    \label{fig:method}
\end{figure*}

%% file: text/3_method.tex
\input{figText/data_samples}
\section{Methodology}
Interactive image editing provides users intuitive operations to flexibly and precisely manipulate real-world images.
In this work, we reformulate it as an image-to-video generation task, and introduce \textit{FramePainter} to inherit real-world dynamic priors from video diffusion models.
To improve fine-grained visual consistency, we propose \textit{matching attention} to encourage each edited image token align with its corresponding source image token.
To train our model, we construct image pairs along with visual editing instructions from high-quality videos.

\subsection{Image Editing as an Image-to-Video Task}
Interactive image editing aims to simulate real-world changes based on user-provided editing instructions, such as sketching, clicking, or dragging. 
Since video corpora provide abundant observations of how the world changes through physical interactions, they can serve as ideal supervision signals for learning such editing behaviors~\cite{Alzayer2024MagicFS,Shi2024LightningDragLF}.
However, existing methods~\cite{Alzayer2024MagicFS,Shi2024LightningDragLF} are typically initialized with pre-trained image diffusion models and lack real-world dynamic priors, thus require a massive number of training samples and additional reference encoders to achieve plausible edits and visual consistency.
In contrast, we reformulate interactive image editing as an image-to-video generation task and introduce \textit{FramePainter} to mitigate above issues.

Fig.~\ref{fig:method} illustrates the pipeline of FramePainter.
Initialized with SVD~\cite{blattmann2023stable}, FramePainter takes a source image $I^\textit{src}$ (\ie, the first frame to SVD) and an editing signal $s$ as input, and then produces a two-frame video that consists of reconstructed and target images.
Inspired by ControlNext~\cite{peng2024controlnext}, we use a lightweight sparse control encoder with multiple ResNet blocks to efficiently encode editing signals.
Notably, we only inject editing signals into target image features to avoid affecting the reconstruction of source image.
The model $\epsilon_\theta$ and control encoder $\phi$ are finetuned with the diffusion loss:
\begin{equation}
\mathcal{L}_\textit{diff} = \mathbb{E}_{\bm{z}_{0}, s, t, \epsilon \sim \mathcal{N}(0, 1)}\Big[ \Vert \epsilon - \epsilon_\theta(\bm{z}_{t},t, I^\textit{src}, \phi(s)) \Vert_{2}^{2}\Big] \, ,
\label{eq:LDM_loss}
\end{equation}
where $\bm{z}_{0}$ is concatenated from source image latent $\bm{z}_{0}^{src}$ and target image latent $\bm{z}_{0}^{tgt}$ along the temporal axis.
Following SVD~\cite{blattmann2023stable}, we use two ways to preserve the visual details of $I^{src}$: (i) inject the features encoded by CLIP image encoder~\cite{radford2021learning} into cross-attention modules, and (ii) 
concatenate $\bm{z}_{0}^{src}$ with per-frame noise latent in channel dimension and then feed them to the denoising U-Net.

\subsection{Matching Attention for Dense Correspondence}
Since SVD relies on 1-D temporal attention to ensure frame consistency, its small receptive field struggles to preserve the identity of objects involving large movements, particularly when only two frames are available (\ie, in image editing task).
Existing works~\cite{khachatryan2023text2video,zhang2023controlvideo} inflate spatial attention into cross-frame attention to enlarge the receptive field, but still fail to achieve fine-grained consistency in appearance~\cite{cong2023flatten,geyer2023tokenflow}.
To alleviate it, we propose matching attention to facilitate dense correspondence between target and source image tokens.

As shown in Fig.~\ref{fig:method}, matching attention is implemented as an auxiliary branch of spatial attention to capture visual details from source image.
Given source image tokens $\bm{z}_{t}^\textit{src} \in \mathbb{R}^{N \times d}$ and target image tokens $\bm{z}_{t}^\textit{tgt} \in \mathbb{R}^{N \times d}$ at timestep $t$, matching attention only takes $\bm{z}_{t}^\textit{tgt}$ as query and $\bm{z}_{t}^\textit{src}$ as key and value:
\begin{align}
    \mathbf{A}^\textit{match} &=\mathrm{Softmax}(\frac{\mathbf{Q}^\textit{match} (\mathbf{K}^\textit{match})^T}{\sqrt{d}}), \\
    \mathbf{O}^\textit{match} &=\mathbf{A}^\textit{match} \cdot \mathbf{V}^\textit{match}, 
    \label{eq:match}
\end{align}
where $\mathbf{Q}^\textit{match} =\mathbf{W}^\textit{match}_Q \cdot\bm{z}_{t}^\textit{tgt}$, $\mathbf{K}_\textit{match} =\mathbf{W}^\textit{match}_K \cdot \bm{z}_{t}^\textit{src}$, and $\mathbf{V}^\textit{match} =\mathbf{V}^\textit{match}_V \cdot \bm{z}_{t}^\textit{src}$.
Next, we pad the output $\mathbf{O}^\textit{match}$ with zeros and add it to the output of spatial attention $\mathbf{O}^\textit{spat}$ as follows:
\begin{align}
    \mathbf{O}^\textit{final} &=\mathbf{O}^\textit{spatial} + [\mathbf{0},\mathbf{O}^\textit{match}], 
    \label{eq:cross_frame}
\end{align}
During training, matching attention copies the weights from spatial attention to inherit its implicit knowledge on image correspondences~\cite{tang2023emergent}.
To encourage more precise correspondence, we employ CoTracker-v3 to extract tracking results, including correspondence matrix $\mathbf{C} \in [0,1]^{N \times N}$ and visible mask $\mathbf{M} \in [0,1]^{N \times N}$: 
\begin{align}
\mathbf{C}[i,j] &= \begin{cases}
1, & \mbox{if } \bm{z}_{t}^\textit{tgt}[i] \mbox{ correponds to }  \bm{z}_{t}^\textit{src}[j]\\
0, &    \mbox{otherwise} \\
\end{cases} \\
\mathbf{M}[i,j] &= \begin{cases}
1, & \mbox{if } \bm{z}_{t}^\textit{tgt}[i] \mbox{ has visible tokens in }  \bm{z}_{t}^\textit{src}\\
0, &    \mbox{otherwise} \\
\end{cases}
\end{align}
Then, we use them optimize the attention weights of matching attention as:
\begin{align}
\mathcal{L}_\textit{match} &= \Vert \mathbf{M} \cdot (\mathbf{A}^\textit{match} - \mathbf{C}) \Vert_{2}^{2}, 
\end{align}
Finally, the overall learning objective is defined as:
\begin{equation}
    \mathcal{L}= \mathcal{L}_\textit{diff} + \lambda_\textit{match} \cdot \mathcal{L}_\textit{match}.
    \label{eqn:final_los}
\end{equation}
where $\lambda_\textit{match}$ controls the scale of matching loss and is set to $1.0$ by default.
As illustrated in Fig.~\ref{fig:main_ablation_attn_vis}, matching attention enables each target token to focus more accurately on its corresponding source token than vanilla cross-frame attention, \ie, translating sparse editing signals into dense correspondences.

\input{tables/main_com}
\subsection{Constructing Samples from Video Data}
Videos capture a wide range of real-world transformations, including object movements, pose variations, and perspective changes, serving as natural supervision signals for image editing task.
Considering the goal of manipulating local regions, we curate high-quality videos with static camera to better support subsequent data construction~\cite{Shi2024LightningDragLF}.
Fig.~\ref{fig:data_samples} presents the examples of various editing signals.

\noindent \textbf{Sampling Suitable Image Pairs.}
To ensure sufficient movements, we randomly sample two frames from the video with a time interval greater than one second.
With the optical flow predicted by SEA-RAFT~\cite{wang2025sea}, we filter out frame pairs whose flow magnitude are excessively small or large, and obtain $22,000$ image pairs in total. 

\noindent \textbf{Extracting Visual Editing Instructions.}
We begin to extract optical flow and tracking points from target image to source image with SEA-RAFT~\cite{wang2025sea} and CoTracker-v3~\cite{karaev2024cotracker3}, respectively.
For each type of editing signal, we design different algorithms to construct them:
\textit{(i)} Sketches.
We utilize a Sobel filter~\cite{zhan2019self} to directly extract the edges of the optical flow, which are then used as the sketch signals.
\textit{(ii)} Dragging points. 
Following LightningDrag~\cite{Shi2024LightningDragLF}, we randomly sample target points with a probability weighted by optical flow magnitude, and then find their corresponding source points according to tracking results.
\textit{(iii)} Coarsely edited images. 
Following Magic Fixup~\cite{Alzayer2024MagicFS}, we employ the softmax splatting algorithm to warp the source image, where the warped images are considered as 
producing a coarsely edited image that aligns with the target.

%% file: figText/data_samples.tex
\begin{figure}[t!]
   \begin{center}
   \includegraphics[width=.99\linewidth]{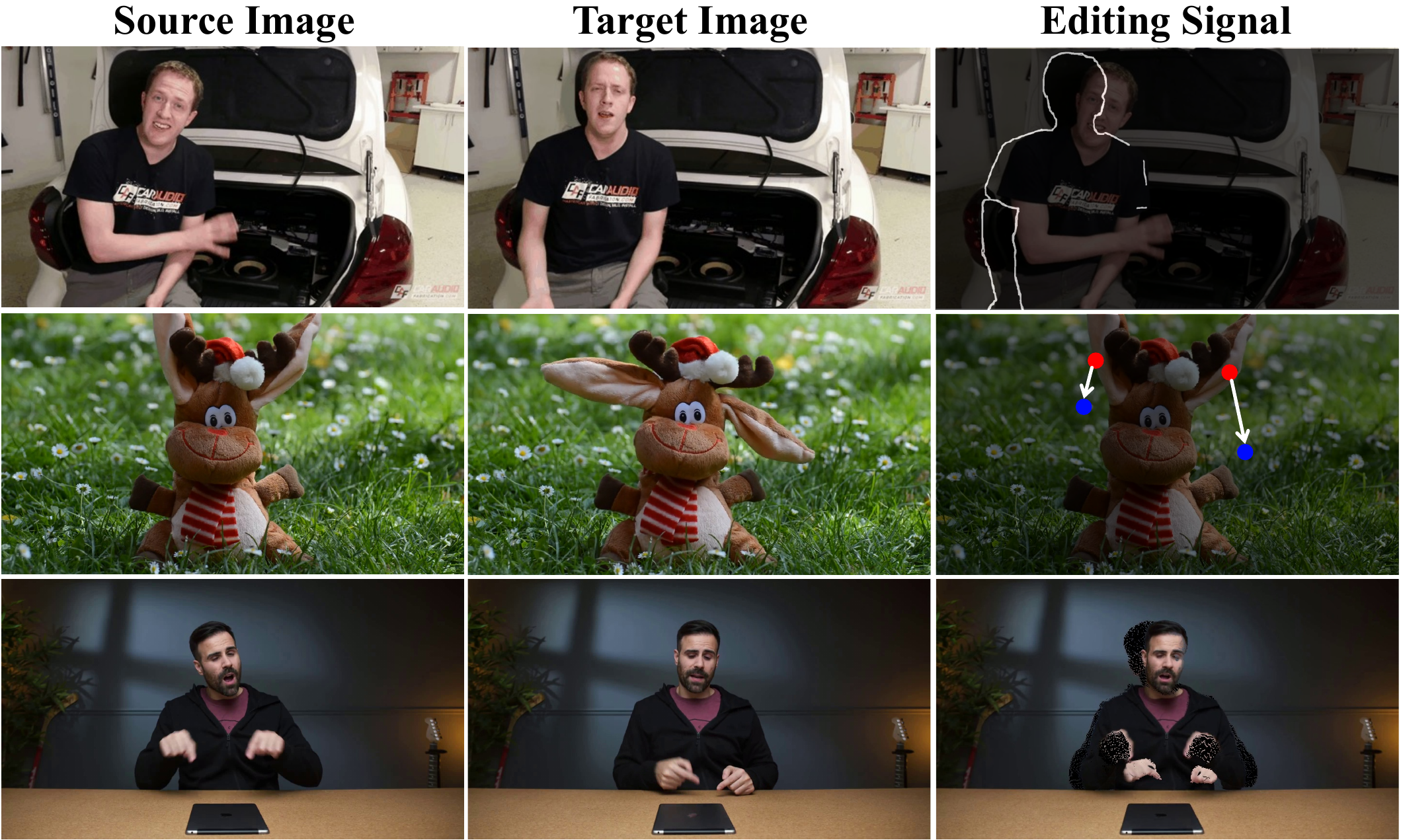}
   \end{center}
   \caption{
   \textbf{Collected samples from videos.}
   We present three types of editing signals from top to bottom: drawing sketches, click points, and dragging regions.
   } 
    \label{fig:data_samples}
\end{figure}

%% file: tables/main_com.tex
\begin{table*}[t]
\caption{
\textbf{Quantitative comparisons across different types of visual editing instructions.}
Despite using fewer than $10\%$ or $1\%$ training samples than previous state-of-the-art methods~\cite{Shi2024LightningDragLF,Alzayer2024MagicFS}, FramePainter surpasses alternative approaches across all editing signals. 
The best results are \textbf{bolded}.
}
\centering
\begin{tabular}{lccccc}
\toprule
Method &Editing Signal &Training Samples & CLIP-FID ($\downarrow$) & LPIPS ($\downarrow$) & SSIM ($\uparrow$)
\\
\midrule
MasaCtrl + ControlNet &Sketch &None &17.933 &0.302 &0.655 \\
\textbf{FramePainter (Ours)} &Sketch &20k &\textbf{7.783} &\textbf{0.140} &\textbf{0.859} \\
\midrule
Magic Fixup &Coarse Edit &2,500k &8.757 &0.166 & 0.855\\
\textbf{FramePainter (Ours)} &Coarse Edit &20k &\textbf{7.573} &\textbf{0.132} &\textbf{0.888} \\
\midrule
DragDiffusion &Points &None &9.192 &0.187 &0.811 \\
LightningDrag &Points &220k &9.894 &0.214 &0.794 \\
\textbf{FramePainter (Ours)} &Points &20k &\textbf{8.513} &\textbf{0.166} &\textbf{0.825} \\
\bottomrule
\end{tabular}
\label{tab:main_com}
\end{table*}

%% file: text/4_experiments.tex
\input{tables/main_user}
\input{figText/main_com}

\section{Experiments}
\subsection{Experimental Setup}
\textbf{Implementation Details.}
We initialize FramePainter with Stable Video Diffusion v1.1 and finetune it on our collected training samples for each editing signal.
Following SVD, the height and width of training images are set to $576$ and $1024$ respectively.
We train the model on two A6000 GPUs with a total batch size of $4$.
The model is optimized for $20,000$ iterations with a learning rate of $1\times 10^{-5}$ using the AdamW algorithm.
During inference, we adopt euler discrete sampling with $25$ steps by default.

\noindent \textbf{Training and Evaluation Datasets.}
We partition the collected samples into training and evaluation datasets, consisting of $\sim 20,000$ and $200$ samples, respectively.
When training the model for each type of editing signal, the only difference is the editing signal used, with the image pairs remaining exactly the same.
Compared to existing evaluation benchmarks~\cite{nie2023blessing}, our evaluation benchmark not only includes source images and editing signals but also provides target images, enabling a more comprehensive assessment of the edited results.

\noindent \textbf{Evaluation Metrics.}
We use three metrics to evaluate the performance of FramePainter:
\textit{(i)} CLIP-FID~\cite{kynkaanniemi2022role} evaluates the semantic alignment between edited images and target images using CLIP embeddings. 
Lower scores indicate better semantic consistency.
\textit{(ii)} LPIPS measures perceptual similarity between images based on deep feature representations. 
Lower values reflect closer visual similarity.
\textit{(iii)} SSIM assesses structural similarity by comparing luminance, contrast, and structure. 
Higher scores indicate better preservation of structural details.

\input{figText/main_application}
\subsection{Comparisons with Baselines}
\noindent \textbf{Qualitative Results.}
Fig.~\ref{fig:main_com} presents the visual comparisons of edited images under different editing signals.
In the top of Fig.~\ref{fig:main_com}, FramePainter not only maintains the visual appearance (\eg, the color of hat), but also achieves more natural and plausible edits than MasaCtrl+ControlNet.
With rich priors from pre-trained video diffusion models, the removal of the car door automatically updates its mirror reflection (highlighted in \textcolor{red}{red box}), ensuring the visual consistency and realism.
In the middle of Fig.~\ref{fig:main_com}, the images refined by FramePainter contain more realistic details, while the results from MagicFixup exhibit noticeable artifacts, \eg, missing head and distorted face.
As shown in the bottom of Fig.~\ref{fig:main_com}, FramePainter excels at maintaining the structural integrity of objects than the alternative baselines.
For example, the baselines produce edited images missing arms or bridge arches, as they focus on keeping appearance with LoRAs or ReferenceNet but overlook structural consistency.

\noindent \textbf{Quantitative Results.}
Table~\ref{tab:main_com} compares the quantitative results in each type of editing signal.
As one can observe, FramePainter significantly outperforms the state-of-the-art approaches under all editing settings, which is consistent with the qualitative results.
In row $1-2$ of Table~\ref{tab:main_com}, it surpasses the performance of MasaCtrl+ControlNet in a large margin.
From row $3$ to $7$, despite using only $1\%$ or even $0.1\%$ of the training data required by previous methods, FramePainter achieves superior performance, highlighting the effectiveness and efficiency of our novel formulation.

\noindent \textbf{User Study.}
To verify the effectiveness of FramePainter, we conduct a user study in $100$ samples consists of source images and editing signals.
We randomly produce $5$ edited images for each sample and obtain $500$ images in total.
For each type of editing signal, we provide the rater a source image, an editing signal, and two edited images from different methods (in random order). 
Then, they are asked to select the better edited image for each of three perspectives: (i) visual consistency, (ii) edit accuracy, and (iii) image quality.
Table~\ref{tab:main_user} summarizes the voting results of raters. 
As one can see, the raters strongly favor the images edited by FramePainter rather than the other baselines from all three perspectives.
\input{figText/main_ablation_attn}
\input{figText/main_ablation_reco}
\input{figText/main_ablation_attn_vis}
\input{tables/main_ablation}
\subsection{Emerging Capabilities of FramePainter}
Despite being trained on only thousands of samples from real-world videos, FramePainter has showcased exceptional generalization across a wide range of scenarios, particularly in out-of-domain contexts.
We provide several representative examples in Fig.~\ref{fig:main_application}.
Firstly, FramePainter supports highly simplified editing instructions and allows users to conveniently modify images, \eg, two sketches are enough to move the antennae or close the eyes in row $1$.
Secondly, it offers intuitive and precise control over complex editing signals, \eg, adjust the shape of the flames and horse tail in row $2$.
Surprisingly, it is also capable of effectively handling out-of-domain scenarios (\ie, cases not found in real-world videos).
For example, in row $3$, users can draw sketches to enlarge the beak of bird or transform the latte art into heart shape.

\subsection{Ablation Study}
\noindent \textbf{Effectiveness of Matching Attention.}
Fig.~\ref{fig:main_ablation_attn} and Table~\ref{tab:ab_attn} compares the performance of FramePainter using different attention mechanisms.
Due to the limitation of receptive field, temporal attention fails to handle editing instructions involving large edited areas (\eg, duplicate mushrooms).
Albeit cross-frame attention can understand editing instructions involving significant changes, it struggles to make each target image token accurately query source image tokens (in Fig.~\ref{fig:main_ablation_attn_vis}), leading to the inconsistency in mushroom color.
In contrast, matching attention improves the accuracy of queries and achieves promising dense correspondence between target and source images (\ie, $3$rd row in Fig..~\ref{fig:main_ablation_attn_vis}).
Therefore, matching attention visibly exceeds both temporal and cross-frame attention in terms of quantitative metrics and visual consistency.

\noindent \textbf{Effectiveness of Source Image Reconstruction.}
Existing image editing methods mainly focus on producing target images with diffusion loss, but ignore source image reconstruction.
Our formulation allows the model to naturally reconstruct both target and source images.
We investigate the impact of source image reconstruction in Fig.~\ref{fig:main_ablation_reco} and Table~\ref{tab:ab_reco}.
From Fig.~\ref{fig:main_ablation_reco}, performing reconstruction improves the visual coherence and realism of edited image (\eg, the color and texture of leaf).
Table~\ref{tab:ab_reco} also highlights the effectiveness of source image reconstruction, \ie, achieves lower CLIP-FID and LPIPS scores.

%% file: tables/main_user.tex
\begin{table}[t]
\caption{
    \textbf{User preference study.}
    The numbers denote the percentage of raters who favor the images edited by FramePainter over other baselines.
    }
    \centering
    \scalebox{0.8}{
    \begin{tabular}{lccc}
    \toprule
         \makecell{Method Comparison} & \makecell{Visual Cons.} & \makecell{Edit Acc.} & \makecell{Image Qual.} \\
         \midrule
         \makecell{Ours vs. MasaCtrl} & $88.3\%$ & $82.0\%$ & $83.2\%$ \\
         \makecell{Ours vs. Magic Fixup} & $71.1\%$ & $72.4\%$ &$76.5\%$\\
         \makecell{Ours vs. LightningDrag} & $73.2\%$ & $68.9\%$ &$72.7\%$\\
    \bottomrule
    \end{tabular}
    }
    \label{tab:main_user}
\end{table}

%% file: figText/main_com.tex
\begin{figure*}[t!]
   \begin{center}
   \includegraphics[width=.95\linewidth]{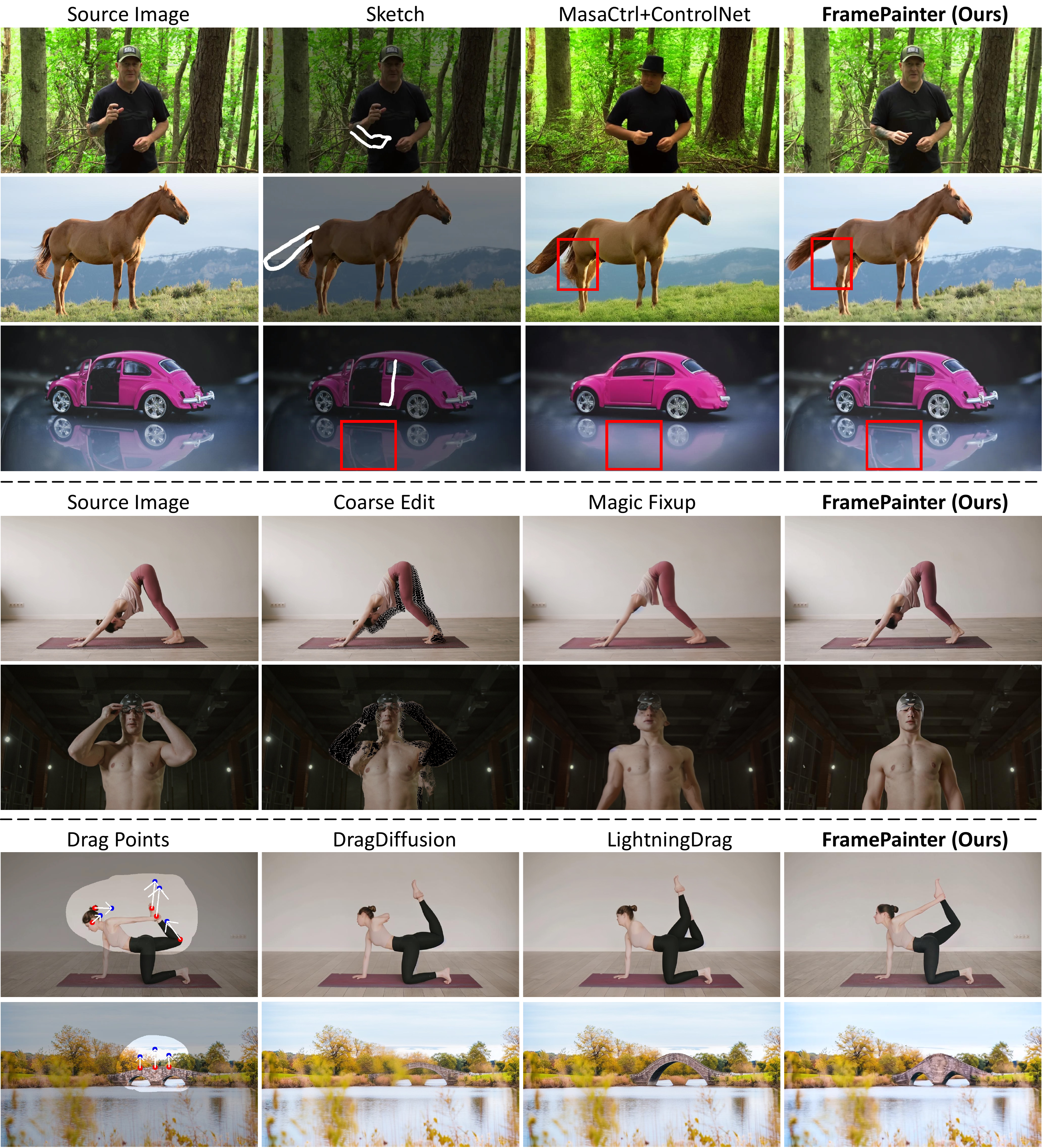}
   \end{center}
   \caption{
   \textbf{Qualitative comparisons across different visual editing instructions.}
   Compared to the baselines, FramePainter not only achieves more coherent and plausible editing results, but also automatically polishes the edited images to meet real-world dynamics, \eg, remove duplicate tail and adjust car door in mirror (highlighted in \textcolor{red}{red box}).
   We note that LightningDrag and DragDiffusion require users to provide additional masks, whereas FramePainter does not.
   } 
    \label{fig:main_com}
    \vspace{-2em}
\end{figure*}

%% file: figText/main_application.tex
\begin{figure*}[ht!]
   \begin{center}
   \includegraphics[width=.95\linewidth]{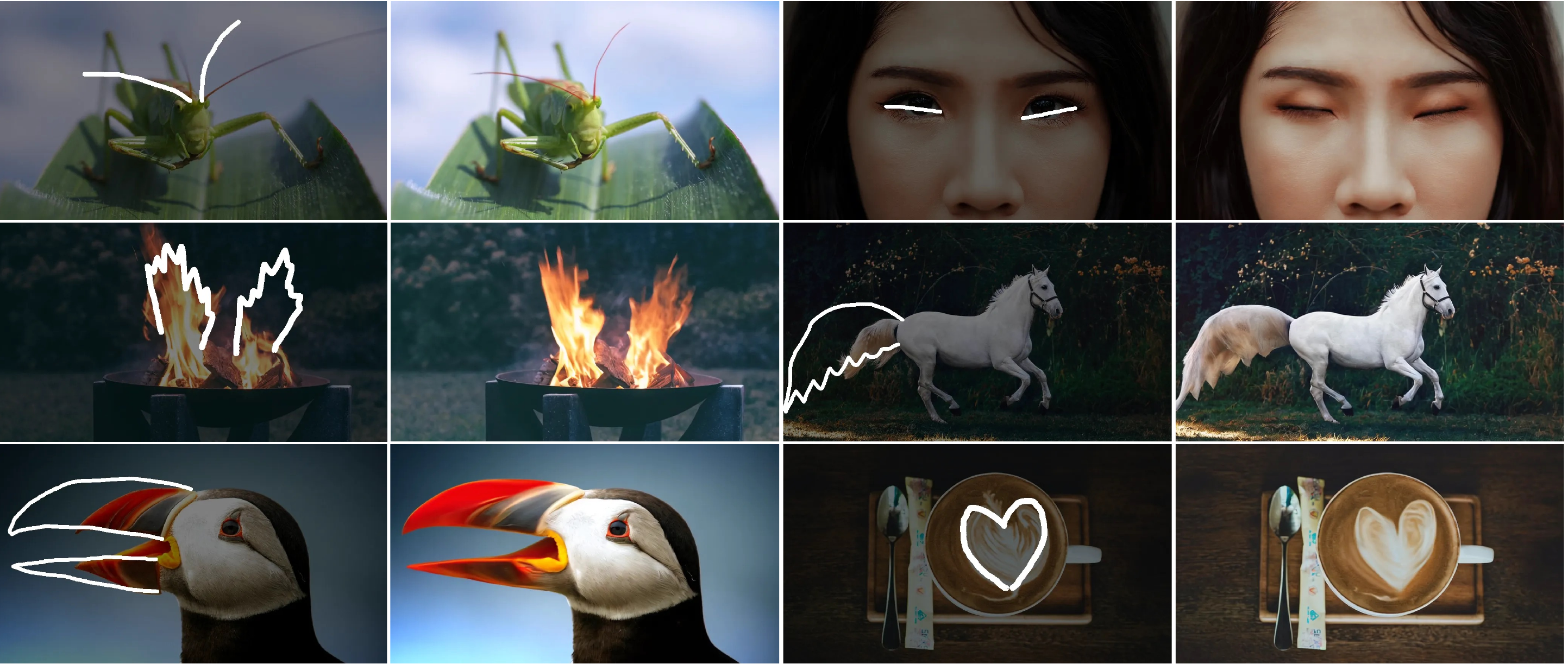}
   \end{center}
   \caption{
   \textbf{Emerging capabilities of FramePainter.}
   Although FramePainter is trained on image pairs from real-world videos, it demonstrates several emerging capabilities as a convenient tool: 
   \textbf{(i)} Supporting highly intuitive and simplified instructions. 
   \textbf{(ii)} Offering precise control over complex editing signals. \textbf{(iii)} Generalizing well to out-of-domain cases, such as shape transformation.
   } 
    \label{fig:main_application}
\end{figure*}

%% file: figText/main_ablation_attn.tex
\begin{figure*}[t!]
   \begin{center}
   \includegraphics[width=.95\linewidth]{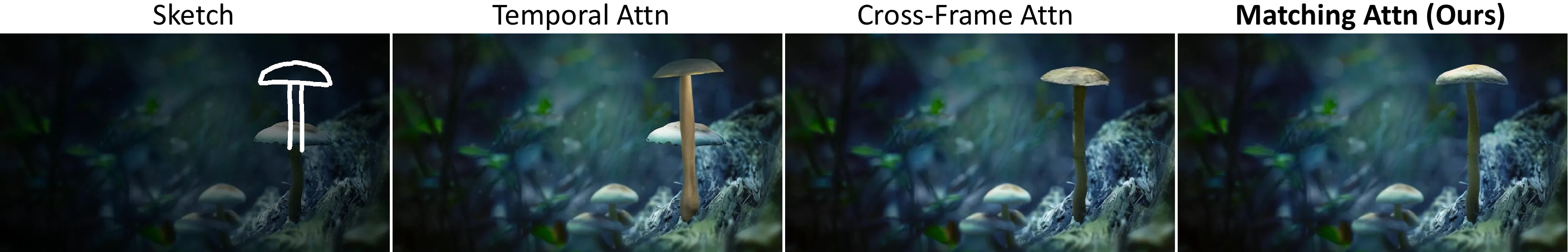}
   \end{center}
    \vspace{-1em}
   \caption{
   \textbf{Qualitative ablation study on the effectiveness of matching attention.}
    Matching attention obtains plausible edited results with fine-grained visual consistency.
    In contrast, temporal attention fails to handle editing signals involving large edited areas, while cross-frame attention struggles to precisely capture appearance.
   } 
    \label{fig:main_ablation_attn}
    \vspace{-1em}
\end{figure*}

%% file: figText/main_ablation_reco.tex
\begin{figure*}[t!]
   \begin{center}
   \includegraphics[width=.95\linewidth]{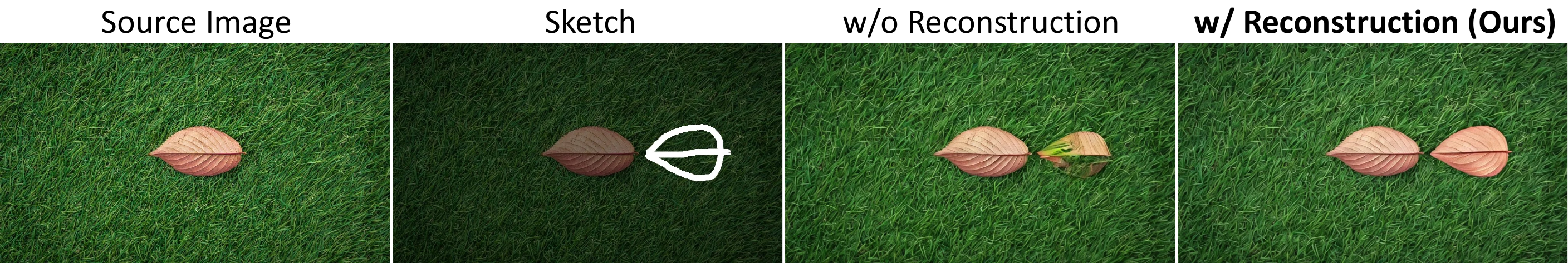}
   \end{center}
    \vspace{-1em}
   \caption{
   \textbf{Qualitative ablation study on source image reconstruction.}
   Compared to w/o reconstruction, reconstruction source image in diffusion loss can better preserve its color and texture and produce more visually consistent edited image.
   } 
    \label{fig:main_ablation_reco}
    \vspace{-1em}
\end{figure*}

%% file: figText/main_ablation_attn_vis.tex
\begin{figure}[t!]
   \begin{center}
   \includegraphics[width=.99\linewidth]{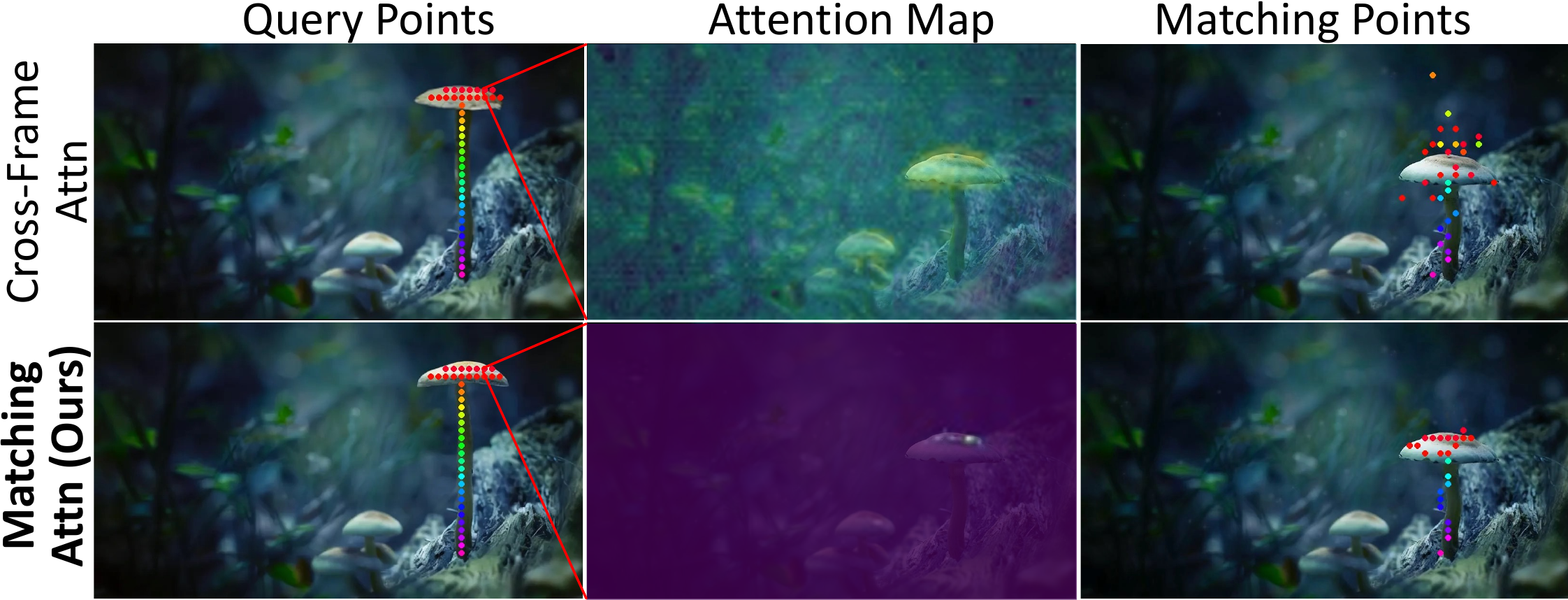}
   \end{center}
    \vspace{-1em}
   \caption{
   \textbf{Visualization of attention weights and dense correspondence.}
   The attention map is computed between the selected target image token (\ie, \textcolor{red}{red} query point) and all source image tokens.
   Among all source image tokens, the token with the highest similarity is marked as the matching point.
   We only visualize the tokens of foreground objects for simplicity.
   } 
    \label{fig:main_ablation_attn_vis}
    \vspace{-1em}
\end{figure}

%% file: tables/main_ablation.tex
\begin{table}[ht]
\caption{
\textbf{Quantitative ablation study on the effectiveness of matching attention.}
Temporal Attn denotes vanilla 1D temporal attention in SVD, while cross-frame Attn is inflated from spatial attention along temporal axis.
}
\centering
\scalebox{0.85}{
\begin{tabular}{lccc}
\toprule
Attention Type & CLIP-FID ($\downarrow$) & LPIPS($\downarrow$) &SSIM ($\uparrow$)
\\
\midrule
Temporal Attn &8.398 &0.165 &0.807 \\
Cross-Frame Attn &8.099 &0.156 &0.826 \\
\textbf{Matching Attn (Ours)} &\textbf{7.783} &\textbf{0.140} &\textbf{0.859} \\
\bottomrule
\end{tabular}
}
\label{tab:ab_attn}
\end{table}

\begin{table}[ht]
\caption{
\textbf{Quantitative ablation study on the effectiveness of source image reconstruction.}
w/ Reconstruction means reconstructing both source and target images in diffusion loss.
}
\centering
\scalebox{0.8}{
\begin{tabular}{lccc}
\toprule
Name & CLIP-FID ($\downarrow$) & LPIPS($\downarrow$) &SSIM ($\uparrow$)
\\
\midrule
w/o Reconstruction &8.201 &0.154 &0.834 \\
\textbf{w/ Reconstruction (Ours)} &\textbf{7.783} &\textbf{0.140} &\textbf{0.859} \\
\bottomrule
\end{tabular}
}
\label{tab:ab_reco}
\end{table}

%% file: text/5_conclusion.tex
\section{Conclusion}
We reframe interactive image editing as an image-to-video generation task, and introduce FramePainter to leverage powerful priors of video diffusion models.
Built upon Stable Video Diffusion, FramePainter greatly reduces training costs while ensuring seamless and coherent image edits. 
Considering the limitations of temporal attention, we propose matching attention to further improve visual consistency by ensuring dense correspondences between the edited and source images. 
The experiments highlights the effectiveness and efficiency of FramePainter, which achieves superior performance than previous state-of-the-art methods with far less training costs.
Additionally, it demonstrates strong generalization to scenarios not present in real-world videos, such as transforming the clownfish into shark-
like shape. 
We hope our work will inspire other image generative tasks that involve priors from videos.

%% file: text/6_appendix.tex
\newpage
\appendix
\onecolumn
\section{Implementation Details of Different Visual Editing Instructions.}
By default, the visual editing instructions (\eg, sketch images and coarsely edited images) are directly encoded using sparse control encoder and injected into the denoising U-Net.
However, it is challenging to encode images that only contain source and target points, which cannot accurately represent the correspondence between each pair of points.
Since this paper aims to explore a general paradigm for interactive learning, rather than focusing on the specific editing method of dragging points, we adopt a simple and intuitive way to encode dragging points.
Specifically, at the output of each attention block, we directly copy the source image tokens corresponding to the positions of source points, and add them to the edited image tokens at the positions of target points.
As a result, this simple approach allows for an accurate understanding of dragging points and enables plausible editing of input images, \eg, in Fig.~\ref{fig:main_com} and Fig.~\ref{fig:sup_drag}.

\section{More Visualizations and Comparisons.}
\begin{figure*}[h]
   \begin{center}
   \includegraphics[width=.99\linewidth]{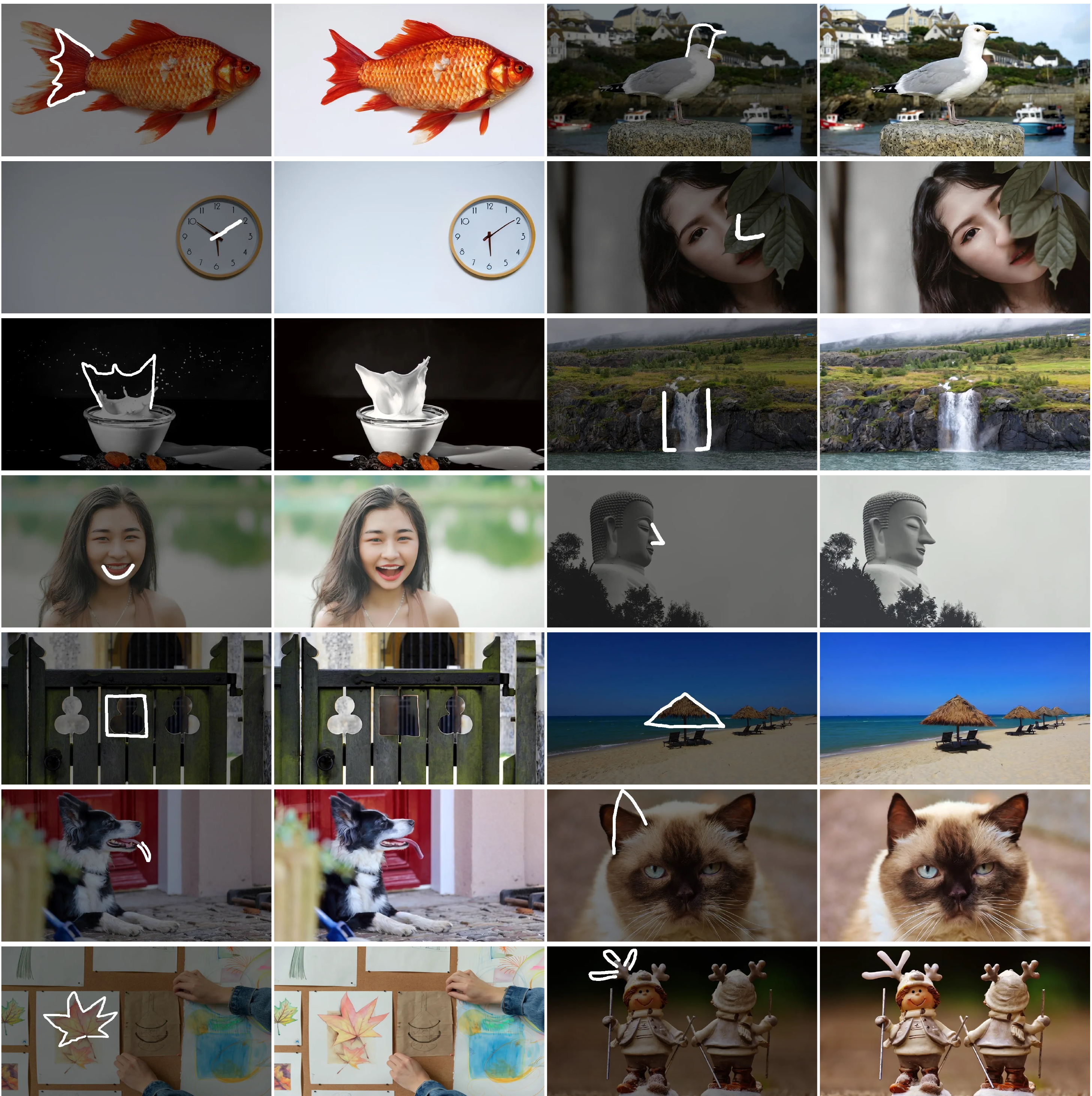}
   \end{center}
   \caption{
   \textbf{More visualization examples of FramePainter.}
   This figure presents both a wide range of scenarios, including in-domain (\eg, change the position of cat ear) and out-of-domain cases (\eg, enlarge the dear horn in hat).
   } 
    \label{fig:sup_sketch_vis}
\end{figure*}

\begin{figure*}[h]
   \begin{center}
   \includegraphics[width=.99\linewidth]{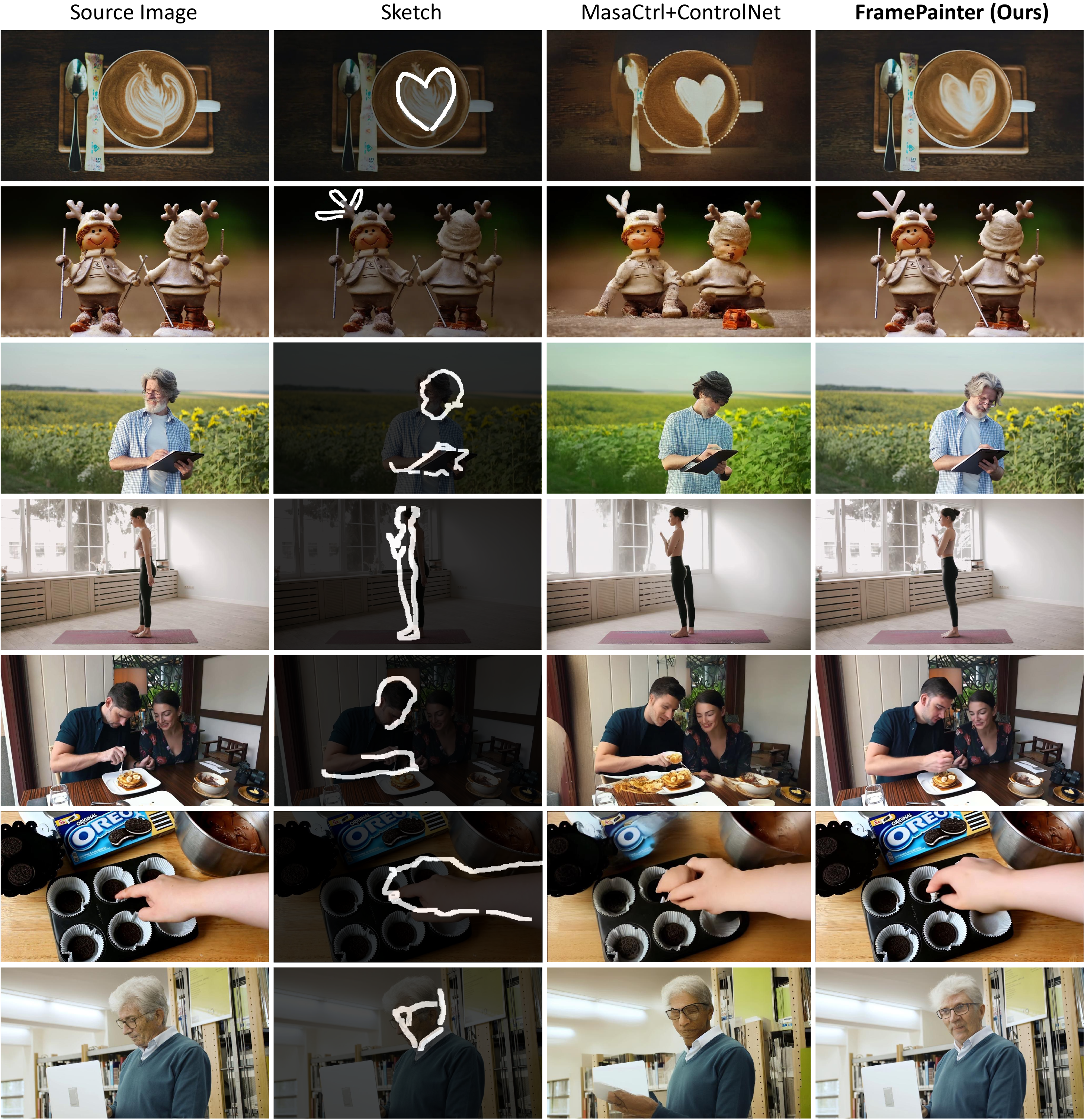}
   \end{center}
   \caption{
   \textbf{More qualitative comparisons in sketch images.}
   } 
    \label{fig:sup_sketch}
\end{figure*}

\begin{figure*}[h]
   \begin{center}
   \includegraphics[width=.99\linewidth]{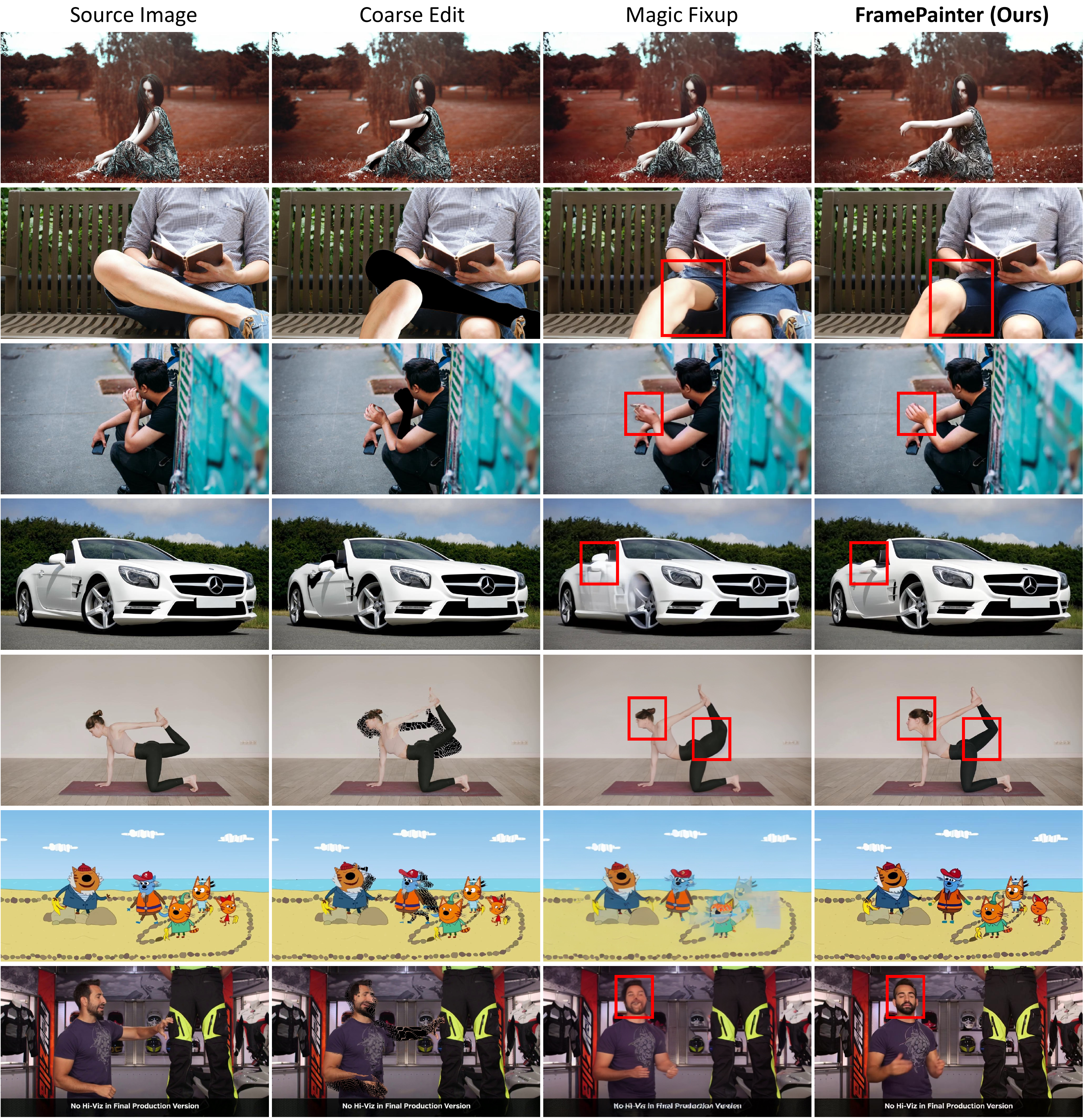}
   \end{center}
   \caption{
   \textbf{More qualitative comparisons in coarsely edited images.}
   } 
    \label{fig:sup_coarse}
\end{figure*}

\begin{figure*}[h]
   \begin{center}
   \includegraphics[width=.99\linewidth]{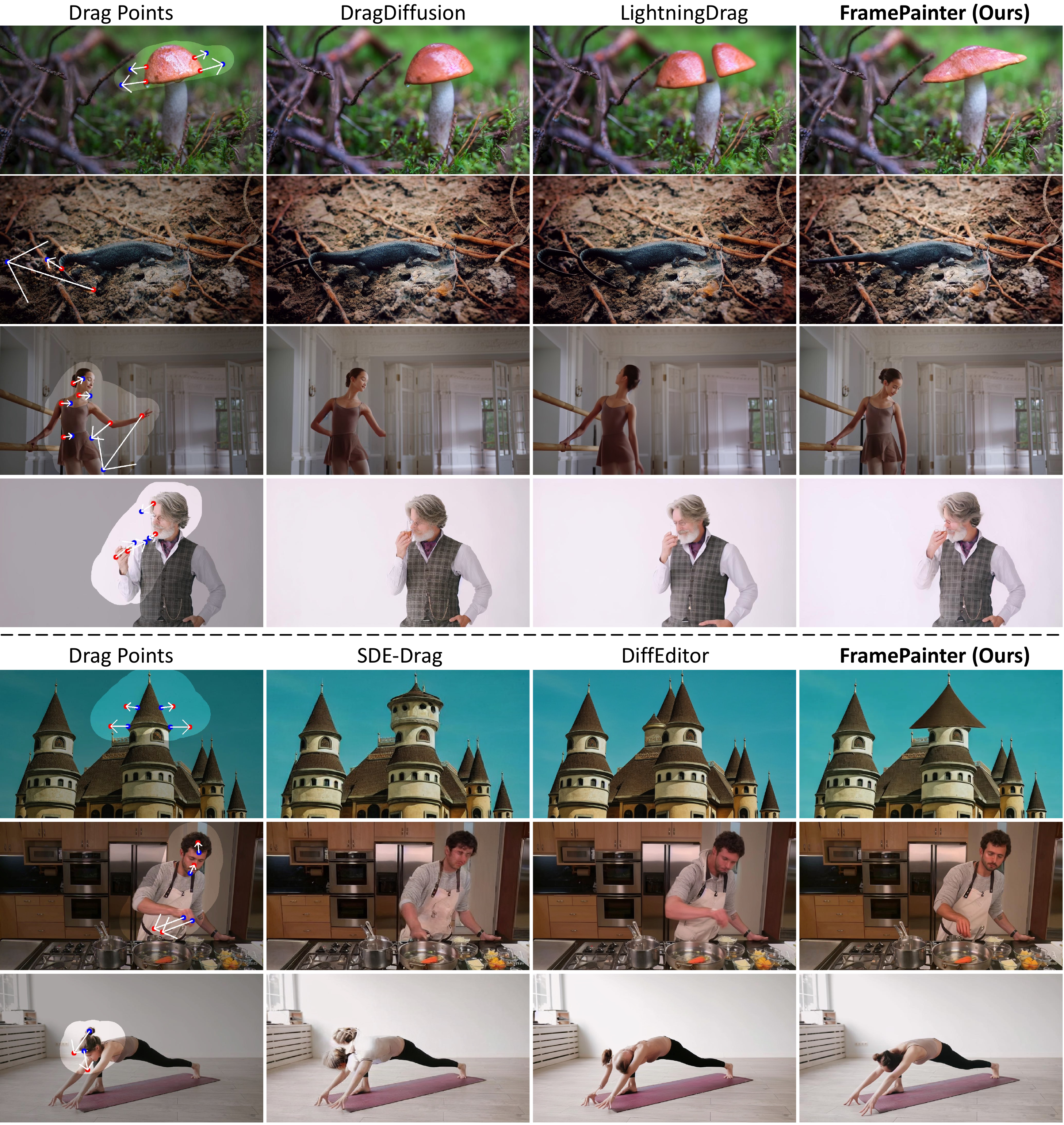}
   \end{center}
   \caption{
   \textbf{More qualitative comparisons in dragging points.}
   We compare FramePainter with both encoder-based (\ie, LightningDrag) and optimization-based methods (\ie, DragDiffusion, SDE-Drag, and DiffEditor).
   During inference, DragDiffusion and SDE-Drag require to finetune additional LoRAs to preserve the visual appearance of source images.
   } 
    \label{fig:sup_drag}
\end{figure*}

Fig.~\ref{fig:sup_sketch_vis} show more visualizations on sketch images.
Fig.~\ref{fig:sup_sketch} and Fig.~\ref{fig:sup_coarse} provide more comparisons with alternative approaches on sketch images and coarsely edited images, respectively.
Fig.~\ref{fig:sup_drag} compares a wide range of drag-based methods, including encoder-based (\ie, LightningDrag~\cite{Shi2024LightningDragLF}) and optimization-based (\ie, DragDiffusion~\cite{Shi2023DragDiffusionHD}, SDE-Drag~\cite{nie2023blessing}, and DiffEditor~\cite{mou2023diffeditor}.
Compared to the baselines, FramePainter presents superior performance in understanding the dragging points and maintaining the structural integrity of objects.
In contrast, due to the absence of real-world dynamic priors, optimization-based methods struggle with moving object parts, \eg, duplicated tail in row $2$ of Fig.~\ref{fig:sup_drag} (top) and duplicate hair in row $2$ of Fig.~\ref{fig:sup_drag} (bottom).
Encoder-based method cannot preserve the overall structure of objects, \eg, separated mushrooms in row $1$ of Fig.~\ref{fig:sup_drag} (top).

